\documentclass[10pt,twocolumn,letterpaper]{article}

\usepackage[pagenumbers]{cvpr} 

\usepackage{graphicx}
\usepackage{amsmath}
\usepackage{amssymb}
\usepackage{booktabs}
\usepackage{float}
\usepackage{multirow}
\usepackage{placeins}
\usepackage{makecell}
\usepackage{rotating}
\usepackage{bm}

\usepackage[pagebackref,breaklinks,colorlinks]{hyperref}

\usepackage[capitalize]{cleveref}
\crefname{section}{Sec.}{Secs.}
\Crefname{section}{Section}{Sections}
\Crefname{table}{Table}{Tables}
\crefname{table}{Tab.}{Tabs.}

\begin{document}

\title{\textit{LookinGood$^\pi$}: Real-time Person-independent Neural Re-rendering for High-quality Human Performance Capture}

\author{
Xiqi Yang$^{1,2}$\ \ \ Kewei Yang$^{2}$\ \ \ Kang Chen$^{2}$\ \ \ Weidong Zhang$^{2}$\ \ \ Weiwei Xu$^{*1}$
\vspace{3pt} \\
\normalsize $^{1}$State Key Lab of CAD\&CG, Zhejiang University \quad \ 
$^{2}$NetEase Games AI Lab \\
\small xiqiyang@zju.edu.cn,\ \  \{yangkewei,\ ckn6763,\ zhangweidong02\}@corp.netease.com,\ \ xww@cad.zju.edu.cn
\vspace{-10pt}
}

\twocolumn[{
    \maketitle
    \begin{figure}[H]
    \hsize=\textwidth
    \centering
    \includegraphics[width=1.0\textwidth]{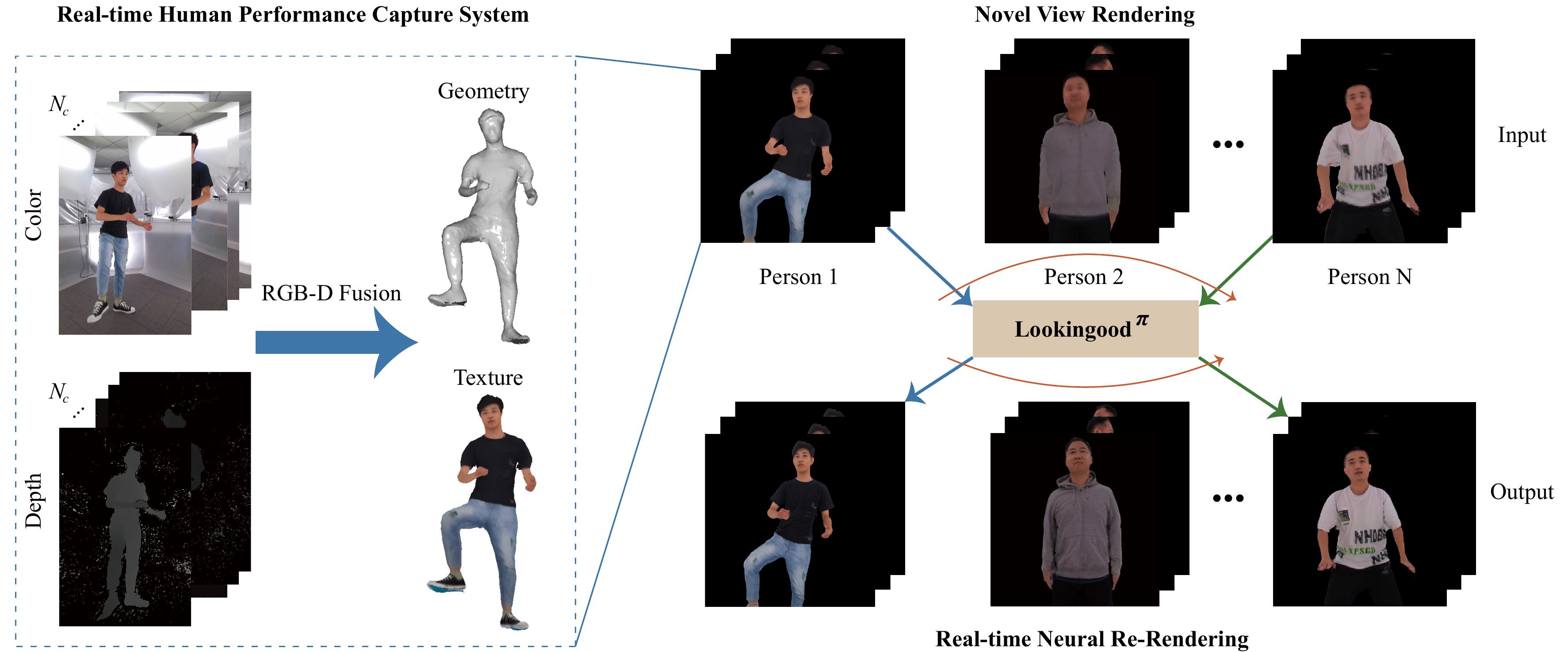}
    \caption{\textit{LookinGood$^\pi$} aims to serve as a general post-processing module to enhance the final rendering quality from a low-cost real-time human performance capture system under sparse multi-view setting (the number of cameras $N_c=8$). Given the reconstructed textured geometry, which contains geometric and texture artifacts, we firstly render it to novel viewpoints and leverage our proposed network to re-render high-fidelity images in real-time. The module is person-independent thus can naturally generalize to novel people.}
    \label{fig:teaser}
    \end{figure}
}]

\begin{abstract}
We propose \textit{LookinGood$^\pi$}, a novel neural re-rendering approach that is aimed to (1) improve the rendering quality of the low-quality reconstructed results from human performance capture system in real-time; (2) improve the generalization ability of the neural rendering network on unseen people. Our key idea is to utilize the rendered image of reconstructed geometry as the guidance to assist the prediction of person-specific details from few reference images, thus enhancing the re-rendered result. In light of this, we design a two-branch network. A coarse branch is designed to fix some artifacts ($i.e.$ holes, noise) and obtain a coarse version of the rendered input, while a detail branch is designed to predict ``correct" details from the warped references. The guidance of the rendered image is realized by blending features from two branches effectively in the training of the detail branch, which improves both the warping accuracy and the details' fidelity. We demonstrate that our method outperforms state-of-the-art methods at producing high-fidelity images on unseen people.

\end{abstract}

\section{Introduction} \label{sec:intro}
Real-time human performance capture has been a popular topic due to the tremendous value in immersive VR and AR applications (\eg telepresence, live performance broadcasting). Existed solutions \cite{Holoportation16,Fusion4D16,Motion2fusion17,livecap19, dynamicfusion15} employ sparse consumer RGB-D sensors setting (even single camera) and volumetric fusion pipeline \cite{kinectfusion11} to generate temporal consistent human models with textures. However, due to the real-time constraint, they often suffer from artifacts in reconstructed geometry and texture (\eg holes, noise, over-smoothed texture). Recently, \textit{LookinGood} \cite{lookingood18} introduces a neural re-rendering approach to enhance human performance capture, which leverages a generative convolutional neural network (CNN) to translate the low-quality rendered images of reconstructed geometry into high-quality rendering results. However, it will suffer from quality degradation on unseen people due to the lack of a generalization mechanism in its network.

To equip the network with a generalization mechanism, NHR \cite{mnhr20} combines a person-specific feature extractor of the reconstructed point-cloud with a shared neural renderer, but with limited generalization performance since it is hard to disentangle person-specific features from the neural renderer totally. Another line of works learn a warping field based on few reference images and corresponding information (\eg keypoints, viewpoints) to synthesize novel views \cite{viewsyn16}, poses \cite{Ren20,fsvid2vid19}, facial expressions\cite{Wang21} \etal. Although they can obtain photorealistic rendering results in many cases, their performances seem to be unstable since the warping accuracy can not be well guaranteed without robust guidance information from the target. In summary, how to improve both the rendering quality of performance capture results and the generalization ability on unseen people is still a challenging problem. 

To address this problem, we propose \textit{LookinGood$^\pi$}, a person-independent neural re-rendering approach, which integrates learning-based image warping into a neural re-rendering pipeline so as to take the advantages of both ($\pi$, namely pi, is the abbreviation of person-independent). Our neural re-rendering pipeline is a two-branch network that can utilize the low-quality rendered image of reconstructed geometry as the approximate guidance to assist the prediction of person-specific details from pre-sampled reference images. Specifically, a {coarse branch} is designed similarly as \cite{lookingood18}, which aims to fix some artifacts in rendered images (\eg holes, noise) and obtain the coarse rendering results, while a {detail branch} is designed to learn a warping field to warp the features of the reference images for the rendering of high-fidelity details. 

The guidance of the rendered image is realized by blending the features from the coarse branch, denoted by guidance features, with the warped features from the reference images in the detail branch. Since the guidance features are learned from the rendered image, these features can encode spatial patterns whose positions are more consistent with the final rendering result. We argue that such information can provide spatially more accurate context information, termed as spatial and appearance guidance hereafter, thus improving the ability of the decoder of the detail branch to correct warping errors, resulting in better rendering quality. In addition, we observe that the gradients obtained with the feature blending scheme are beneficial to improve the accuracy of the learned warping field.

We demonstrate better generalization performance under more sparse cameras setting than state-of-the-arts. We also show that our approach can run in real-time and preserve high-quality rendering results under novel view, which makes it available for online VR/AR applications.

We summarize our contributions as follows:
\begin{itemize}
    \item We present \textit{LookinGood$^\pi$}, a real-time person-independent neural re-rendering approach, to enhance human performance capture, especially with sparse multi-view setting. 
    \item We propose a novel feature blending scheme to combine information from two branches effectively in the training of the detail branch, improving both the warping accuracy and the details' fidelity.
    \item We demonstrate the state-of-the-art performance at producing high-fidelity images on unseen people.
\end{itemize}

\section{Related work} \label{sec:relatedwork}
In this section, we review the researches mostly related to our work, including human performance capture, neural re-rendering, and learning-based image warping.     

\textbf{Human performance capture.} Marker-less human performance capture systems \cite{Edilson08, Christoph98} have been widely used to reconstruct the geometry of dynamic human and generate the free-viewpoint video, which can be briefly divided into two categories: offline and real-time systems.
Offline systems\cite{totalcapture18,fvv15,relightable19} usually rely on a well-controlled multi-camera studio and a huge amount of computational resources to achieve high-quality reconstruction and rendering results. Recently, real-time systems have been proposed to open-up online applications such as telepresence\cite{Holoportation16}, which aim to combine non-rigid tracking\cite{li09,Sumner07} and volumetric fusion\cite{Curless96} to generate temporally consistent human models. Since real-time single-view solutions often suffer tracking failure due to large topological changes or complex motions \cite{dynamicfusion15, RobustFusion20}, we turn to the lightweight multi-view solutions \cite{Fusion4D16, Motion2fusion17}, which only rely on few RGB-D sensors (from 3 to 8) to obtain compelling volumetric capture. However, multi-view solutions always suffer from both geometric and texture artifacts due to inconsistent lighting conditions and calibration imprecision across views, which can not satisfy the quality requirement for immersive VR/AR experiences.

\textbf{Neural re-rendering.}  A widely-used approach in neural re-rendering is to collect a large amount of training data per-scene/subject to train the rendering network~\cite{dnr19,npbg20,nrwild19,fsnhr21}. However, it is difficult to apply the approach for online re-rendering for objects that are not in the training data. Several works~\cite{VolumetricCapture19, NeuralHumanFVV21} claim generalization ability on novel subjects. However, they focus on synthesizing novel images by blending few neighboring images with a coarse geometry, which may lead to artifacts in occluded regions. For the human re-rendering, it is also possible to learn the appearance features at 3D vertices from images to directly re-render the reconstructed parametric 3D human models, such as SMPL or SMPL-X~\cite{StylePeople21,SMPLpix21,SMPL15,SMPLX19}, which is usually time-consuming to train for the vertex-level appearance code. The works mostly similar to us are \textit{LookinGood} \cite{lookingood18} and NHR \cite{mnhr20}, but with limited generalization ability to achieve high-quality rendering results for unseen subjects. Different from them, we focus on refining rendered images using reference images and propose a guidance scheme to improve the warping accuracy and the rendering quality. 

\textbf{Learning-based image warping.} CNN has been applied to learn warping fields from image features to transfer the appearance of the subject in reference images to novel view/pose/facial expressions~\cite{Neverova18, Yining19, Grigorev19,Ren20,Wang21}. For example, \cite{firstorder19} and \cite{monkeynet19} estimate a warping field based on 2D body key-points to synthesize images under target poses. \cite{fsvid2vid19} further introduces an attention module to select a reference image with the pose most similar to the target, which makes warping easier. These methods can produce high-fidelity results in many cases if the estimated warping field is accurate. However, it is challenging to guarantee the warping accuracy due to occlusions, lighting changes in the images. In contrast, we introduce guidance information from the rendered image to significantly improve the warping accuracy for 3D human re-rendering. 

\section{Method}
\begin{figure*}
  \centering
  \includegraphics[width=1.0\textwidth]{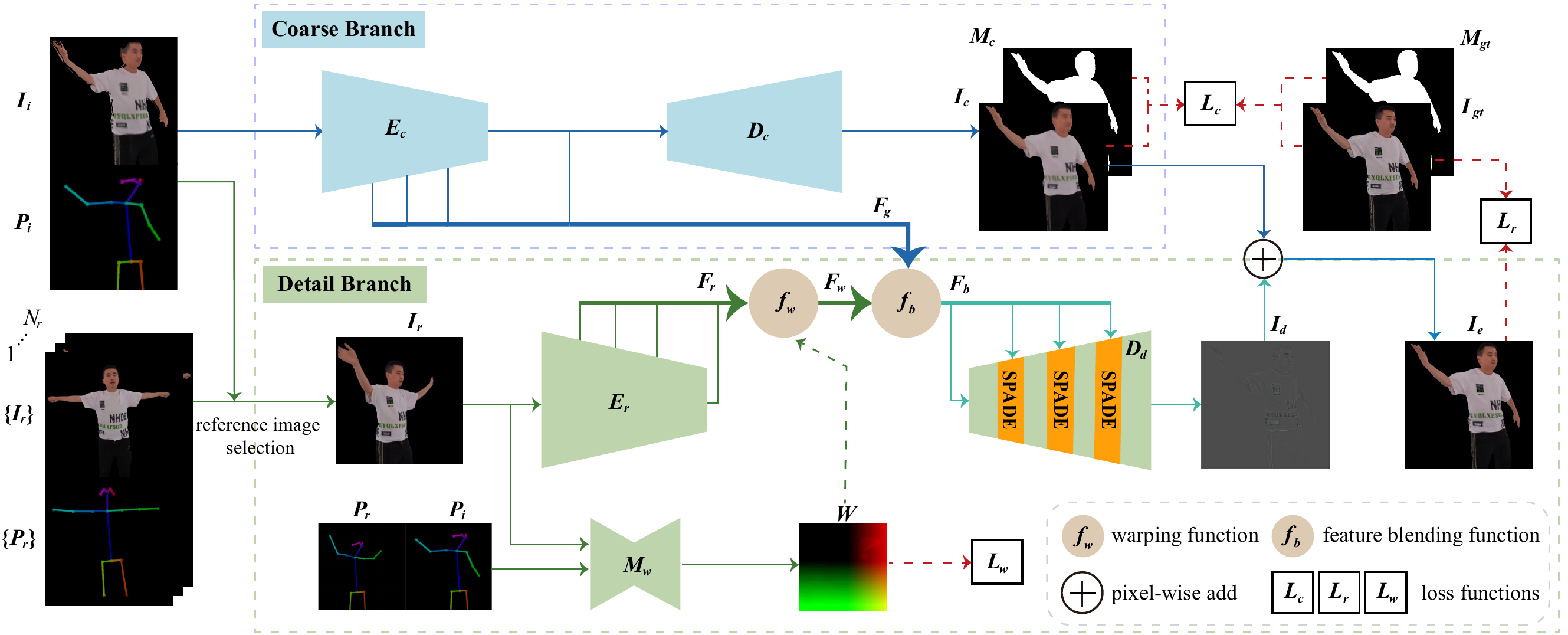}
  \caption{Overview of our proposed two-branch network. The coarse branch aims to fix geometric artifacts in the rendered image $I_i$ of the reconstructed geometry and outputs a coarse image $I_c$ and a binary segmentation mask $M_c$, while the detail branch predicts a person-specific details image $I_d$ from a selected reference image $I_r$, which is combined with $I_c$ to obtain the final enhancing result $I_e$. The blending function is used to introduce the guidance information thus guiding the detail branch to predict ``correct" details. We supervise the whole network using three losses: the coarse loss $L_c$, the reconstruction loss $L_r$ and the warp loss $L_w$.}
  \label{fig:overview}
\end{figure*}

The overview of our system is shown in \cref{fig:overview}. We adopt a multi-view setting, specifically, we build a 360-degree human performance capture system using $N_c$ calibrated cameras ($N_c=8$ in our case). Given a synchronized multi-view RGB-D video sequence, we leverage the state-of-the-art real-time volumetric reconstruction and texturing methods \cite{doublefusion18, albedo17} to obtain an approximate per-frame textured geometry, which can be rendered in arbitrary view. We further capture $N_r$ images for each performer with specific pose (\eg T-pose, A-pose) from $N_c$ views as reference images. 

Our proposed network is composed of two branches: (1) the coarse branch learns a coarse image $I_{c}$ and a foreground mask $M_c$ from the rendered input image $I_i$, (2) the detail branch predicts a person-specific details image $I_d$ from a selected reference image $I_{r}$, and generates the final enhancing image $I_e$ by combining $I_c$ and $I_d$. In the following, we will illustrate the architectural design of our proposed network as well as the training strategy in more detail. 

\subsection{Coarse Branch}
The coarse branch learns to (1) extract useful patterns in the rendered image $I_i$ as the guidance; (2) inpaint the missing texture regions due to imprecise reconstructed geometry, thus obtaining a coarse image $I_c$. Note that the coarse branch is supposed to be lightweight with a small capacity since we found that complex models will lead to the gradual dominance of the coarse branch during training and eventually disable the detail branch. This hinders the network to decompose coarse and detail information effectively, leading to degeneration when applied to novel people.

Considering the effectiveness and real-time requirement, we resort to a U-Net like architecture used in \cite{lookingood18} but make some modifications to better fit with our system. For example, we reduce the layer numbers in both encoder $E_c$ and decoder $D_c$ and modify the downsampling method. To be better applied to VR/AR rendering, we predict a binary segmentation mask $M_c$ as a byproduct, which indicates foreground pixels in $I_c$. Hence the final layer of $D_c$ outputs a four-channel image, including $I_c$ and $M_c$. Note that we use the downsampled rendered image (by two times) as input and upsample results to the original resolution as output.

In addition to the final coarse output, the coarse branch also provides multi-layer encoded features, which we call guidance features, denoted by $F_g$. $F_g$ encodes multi-level coarse information in $I_i$, which can serve as a rough spatial and appearance guidance for the detail branch.

\subsection{Detail Branch}
The detail branch learns to (1) extract multi-level features $F_r$ in reference images $\{I_r^i\}_{i=1}^{i=N_r}$ by a reference encoder $E_r$ (\cref{sec:reference_encoder}); (2) select a ``best" reference image and estimate a warping field by a warp module $M_w$ based on the pose representation of $I_i$ and selected $I_r$ ($i.e.$ sparse 2D body keypoints, denoted by $P_i$ and $P_r$), which is used to warp the $F_r$ to the input pose (\cref{sec:warp_module}); (3) combine the warped features $F_w$ with the guidance features $F_g$ to generate a detail image $I_d$ by a detail decoder $D_d$ (\cref{sec:detail_decoder}). 

\subsubsection{Reference encoder $E_r$} \label{sec:reference_encoder}
For each performer, we pre-capture some images with specific poses (\eg T-pose, A-pose) as reference images. Since these images contain a large amount of person-specific texture information, to better extract useful patterns, we employ a 4-convolutional-layer encoder $E_r$ like most works \cite{mustgan21,fsvid2vid19} to extract features for each reference image $I_r$ at different levels, denoted by $F_r$. 

\subsubsection{Warp module $M_w$} \label{sec:warp_module}
Given multiple encoded reference images, an alternative scheme is feeding all of them to the network, learning multiple warping fields and blending them to obtain the final warped result. Although it may bring more information to the result, the blending process will lead to blurring around boundaries (as illustrated in \cite{rife20}), which is against our goal of predicting high-fidelity details. Thus, we only select a reference image from all candidates, which follows two principles: (1) the mean Euclidean distance between two roughly aligned poses $(P_i, P_r)$ should be as small as possible; (2) the number of corresponding feature points (\eg SURF) in two images $(I_i, I_r)$ should be as large as possible. Moreover, when calculating similarity scores, we penalize the missing keypoints since occlusion may occur.

Then, similar to \cite{monkeynet19}, we learn a pixel-wise warping field $W$ based on $P_i$, $P_r$ and $I_r$. $W$ is composed of two parts: (1) a coarse field $W_c$ to estimate part-based rigid motion; (2) a refine field $W_r$ to predict local non-rigid motion, which needs to be trained carefully, see Supplemental Materials for more details. $W$ is applied to multi-level reference features $F_r$ to obtain the warped features: $F_w=f_w(F_r)$, where $f_w$ refers to the warping function. We use downsampled images (by four times) for the warping field learning and upsample to desired resolutions when applying.

\subsubsection{Detail decoder $D_d$} \label{sec:detail_decoder}
\begin{figure}
  \centering
  \includegraphics[width=1.0\linewidth]{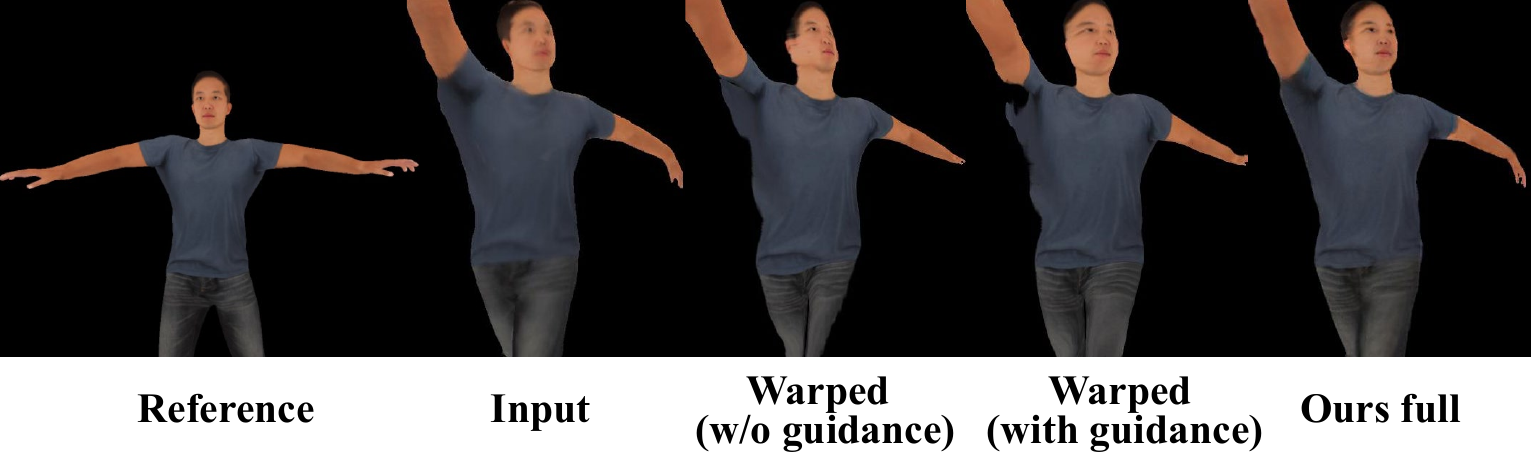}
  \caption{Our method extracts the spatial and appearance information from the rendered input image as a guidance, which not only improves the warping accuracy (from the reference to the input image) but also guides our detail branch to predict ``correct" details from the warped result. Zoom-in for the best of view.}
  \label{fig:warping}
\end{figure}

As the core of the detail branch, the detail decoder $D_d$ aims to learn a ``correct" detail image $I_d$ by combining $F_g$ and $F_w$ effectively. After combination, the spatial and appearance guidance information embedded in $F_g$ can be effectively preserved and propagated throughout the decoder and the warping field, which will contribute to improving the warping accuracy as well as guide $D_d$ to learn \textit{``where to extract correct details"}, as illustrated in \cref{fig:warping}.

To combine $F_w$ and $F_g$, a naive strategy is concatenating them and feeding them as input to the decoder, which will not only increase the model size but also make it more difficult to learn useful information in $F_g$ and $F_w$ due to the larger optimization space. In this paper, we propose to blend them and serve them as the conditional input to the decoder. Specifically, we introduce a hyper-parameter called $\alpha$ to control the blending process and calculate the blending feature $F_b$ for each layer as \cref{eqn:blending}. It is desirable to develop a module to learn the adaptive $\alpha$ values according to the input. However, we found it was hard to figure out a simple yet efficient way to learn features that are correlated to the $\alpha$ during our experiments. Currently, to ensure that the re-rendering pipeline is fast enough, we choose to determine the $\alpha$ manually through experiments, see \cref{sec:ablation} for more details. We leave the task of designing a scheme to learn spatially-variant $\alpha$ as a future work. 
\begin{equation}
    F_b = F_g*\alpha + F_w*(1-\alpha) \label{eqn:blending}
\end{equation}

Given the multi-layer blending features $F_b$, we take the last-layer feature as the input to $D_d$, and remaining-layer features as the conditional input to learn the spatial modulating parameters like \cite{spade19}. The decoder is designed like the SPADE generator used in \cite{spade19} with minor modifications, \eg we add additional convolutional layers to enlarge the model capability and remove the residual blocks. 
 
\subsection{Network training} \label{sec:trainingstrategy}

To train the whole network more effectively, we divided the training into two stages: a coarse branch training stage and a detail branch training stage. Although our network can directly generalize to novel people after training, we introduce a fine-tuning stage to further improve the fidelity of generated images by using few data from novel people. See \cref{sec:imp_details} for more details about training settings.

\textbf{Coarse branch training stage.}
In this stage, we train the coarse branch by supervising the coarse loss $L_{c} = |I_{c}-I_{gt}| + |M_{c}-M_{gt}|$, which measures the quality of the coarse image and the predicted mask. Since it is unnecessary to generate a high-fidelity image for this branch, we choose $l_1$ loss instead of perceptual loss.

\textbf{Detail branch training stage.}
In this stage, we fix the weights of the coarse branch and train the detail branch by supervising the reconstruction loss and the warp loss.

\textit{Reconstruction Loss $L_{r}$} measures the distance of the final enhancing image $I_{e}$ and the groundtruth image $I_{gt}$. To improve the image quality, we adopt a perceptual loss term $L_{r}^{vgg}$. Moreover, we add a small $l_1$ loss term $L_{r}^{img}$ to speed-up convergence. The loss is defined as \cref{eqn:rec_loss}:
\begin{align}\label{eqn:rec_loss}
    L_{r}^{vgg} = \sum_i|V&GG_i(I_{e})-VGG_i(I_{gt})|
    \notag
    \\
    L_{r}^{img}& = {|I_{e}-I_{gt}|}
\end{align}
where $VGG_i(\cdot)$ denotes the $i^{th}$ feature extracted from the pretrained VGG-19 layer. We follow \cite{firstorder19} to calculate this loss in multiple scales.

\textit{Warp Loss $L_{w}$} measures the warping accuracy. Firstly, we calculate difference between the warped reference image and the groundtruth, namely $L_{w}^{img} = |f_w(I_r)-I_{gt}|$, where $f_w$ denotes the warp operation. As described in \cref{sec:warp_module}, the learned warp field $W$ is composed of two parts: the coarse field $W_{c}$ and the refine field $W_{r}$. Since $W_{r}$ is a per-pixel motion field, we further add a $l_1$ regular term $L_{w}^{reg} = |W_{r}|$ to control its value thus avoiding wrong gradient descent direction at the beginning. 

In summary, we formulate the total loss $L_d$ as \cref{eqn:total_loss}. The loss weights $(\lambda_{r}^{vgg},\lambda_{r}^{img},\lambda_{w}^{img},\lambda_{w}^{reg})$ are set to $(0.9,0.1,1.0,1.0)$ to ensure that all losses provide a similar contribution. Note that we only set $\lambda_{w}^{reg}=1.0$ in the early training phase, otherwise set it to 0.
\begin{align} \label{eqn:total_loss}
    L_d &= \lambda_{r}^{vgg}*L_{r}^{vgg} + \lambda_{r}^{img}*L_{r}^{img} 
    \notag 
    \\&+ \lambda_{w}^{img}*L_{w}^{img} + \lambda_{w}^{reg}*L_{w}^{reg}
\end{align}

\textbf{Fine-tuning stage.} In this stage, we fix the weights of $E_r$ and fine-tune other modules since we found it results better. We combine the losses used in two training stages mentioned above. Namely, the fine-tuning loss $L_f=\lambda_c*L_c+\lambda_d*L_d$, where $\lambda_c=0.5$ and $\lambda_d=1.0$.

\section{Experiments}
In this section, we evaluate the performance of our proposed framework against several state-of-the-art methods on our datasets. Then, we analyze the neural re-rendering time of our method. Furthermore, we perform some ablation studies to verify the effectiveness of our network.

\subsection{Datasets}
Our dataset contains captured data and synthetic data. To prepare captured data, we placed 8 pre-calibrated and synchronized Azure Kinect RGB-D sensors evenly distributed around the performer to record multi-view RGB-D sequences. Then we implemented a real-time volumetric reconstruction system like DoubleFusion framework\cite{doublefusion18} and \cite{albedo17} to obtain the reconstructed textured geometry. We perform segmentation on the captured images by using the state-of-the-art method \cite{MGM21} and obtain the groundtruth foreground masks, which are further used to crop and resize all images from $1280\times720$ to $1024\times512$. To obtain body keypoints, we optionally select a popular real-time pose detection API: Openpose \cite{openpose21}.
Since captured data is relative hard to collect, we use Renderpeople\cite{renderpeople} to build our synthetic dataset. Specifically, we drive selected rigged subjects using motions from Mixamo\cite{mixamo} and record 8-view RGB-D sequences as the input of our reconstruction system. To simulate the captured data better, we add noise with Kinect noise patterns \cite{kinectnoise15} to depth images, and perturb the camera parameters slightly to simulate the calibration error.

The constructed dataset comprises 17 performers (4 captured and 13 synthetic) performing different actions in different clothing. For each performer, we captured some reference images with specific poses and recorded 8 sequences of about 150 frames. We select 9 synthetic subjects for training and left 64 test sequences of remaining subjects (4 synthetic and 4 captured) for evaluation.

\subsection{Implementation details} \label{sec:imp_details}
We use Pytorch to implement our proposed framework and adopt Adam optimizer with weight decay to avoid overfitting during the training and fine-tuning stage. The learning rate is initialized to $5e-5$ and the weight decay factor is $3e-6$. We train the network on a single NVIDIA V100 GPU for around 50K iterations with a batch size of 4, which takes about 50 hours to converge. 

As for fine-tuning on novel people, we only need to randomly select 5 frames (8 views) from a pre-captured short sequence as input, and 4 frames (8 views) from the reference set of the same person as references ($i.e. N_r=32$), then run fine-tuning for 20 epochs, which takes about 5 minutes. The value of $N_r$ will be further discussed in \cref{sec:ablation}. 

\subsection{Neural re-rendering time analysis}
We analyze the re-rendering time of our approach on a single NVIDIA V100, with the $1024\times512$ resolution of both the rendered input and output image. To speed-up model inference, we cast the model weights to 16-bit precision and we do not observe any obvious quality degradation. 

The average re-rendering time is $41ms$, with $4ms$ for coarse estimation, $7ms$ for reference encoding, $11ms$ for warping, and $19ms$ for detail decoding. Since the reference encoding only needs to be calculated once for each person, the online time can be decreased to $34ms$ ($i.e.$ 29fps), which obviously satisfies the real-time requirement.

\subsection{Comparison with state-of-the-art methods}

In the following, we compare the performance of our approach with two state-of-the-art methods: \textit{LookinGood}\cite{lookingood18} and NHR\cite{mnhr20}, which are also aim to produce a high-quality rendering result from a low-quality human reconstruction. We implement \textit{LookinGood} as described in their paper and retrain \textit{LookinGood} and NHR from scratch using the same data as ours (for NHR we use the per-frame reconstructed textured point-cloud instead of the rendered image as input).

We perform the comparison experiments from two perspectives. Firstly, we evaluate the image-enhancing performance on test sequences qualitatively and quantitatively, of which the groundtruth is available. Secondly, we evaluate the robustness under novel view. Note that we conduct all experiments on \textit{unseen} subjects since our motivation is to design a person-independent framework, which means it is supposed to achieve better generalization performance. We provide additional results in Supplementary Materials.

\textbf{Evaluation on test sequences.} \label{sec:eval_testseq}
We choose 8 unseen subjects from our datasets (captured dataset and synthetic dataset), for which we select a test sequence (about 50 to 100 frames, 8 views) to perform evaluation. To ensure fairness, we fine-tune \textit{LookinGood} and NHR as described in their paper, using the same data and same iterations as ours. Then we run the quantitative and qualitative evaluation for remaining frames. We adopt multiple standard metrics used in the image-to-image translation tasks \cite{lookingood18, mnhr20} to measure the difference between the predicted image and the groundtruth: PSNR \cite{psnr10}, SSIM \cite{ssim04}, MSE, LPIPS \cite{LPIPS18}.

\begin{figure*}
  \centering
  \includegraphics[width=1.0\textwidth]{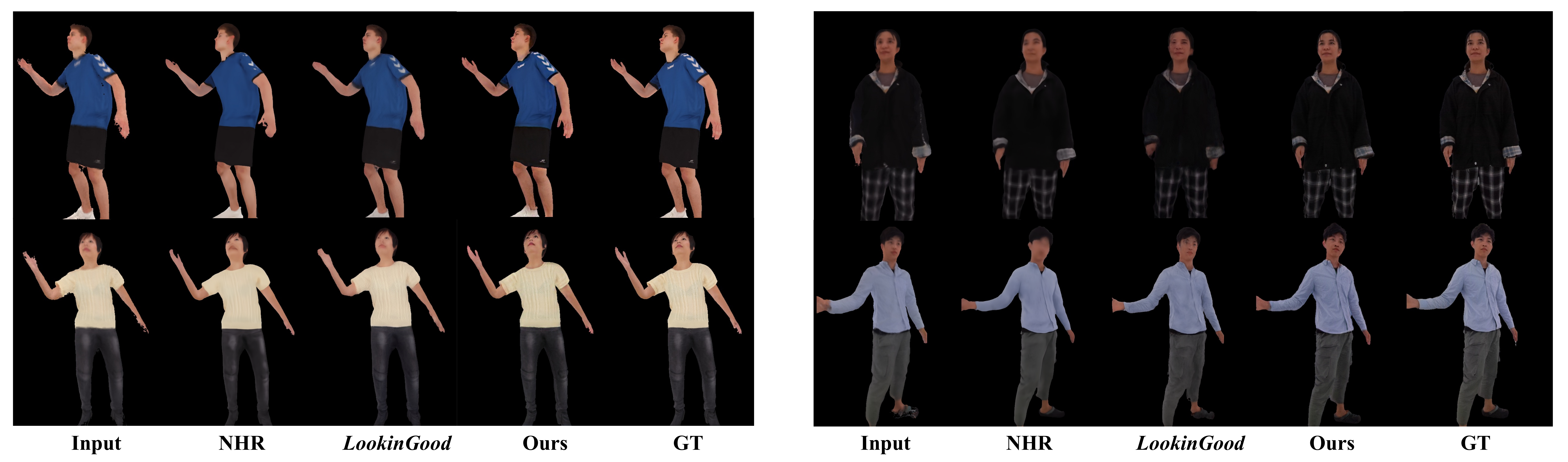}
  \caption{Results on test sequence of novel people. Our method can not only fix the missing texture areas in the input image due to geometric artifacts but also translate the over-smoothed texture to a higher-quality rendering, which achieves the best generalization performance than others. Zoom-in for the best of view.}
  \label{fig:test_seq}
\end{figure*}

As shown in \cref{tab:test_seq}, our method outperforms other methods on all metrics. We also present qualitative results in \cref{fig:test_seq}. Although NHR can restore some useful appearance information due to its generalization mechanism, the results are still unsatisfactory, which demonstrates that it can not disentangle the person-specific point-cloud feature extractor and the shared neural renderer very well in the case of less training data, especially when the point-cloud is sparse and noisy (as described in their paper). \textit{LookinGood} improves the rendering quality to some extent, but its result loses a large amount of appearance details and looks a little blurry, which showcases the limited capability of a separate CNN to generalize to novel subjects. In contrast, our proposed method achieves the best generalization performance on unseen subjects due to our effective two-branch design. 

Note that we also found that the generalization performance of NHR and \textit{LookinGood} is highly-related to the data size used in fine-tuning. Since our goal is to obtain satisfactory results by using less training data, we only select 5 frames (8 views) for fine-tuning. We will evaluate more fine-tuning data size settings in Supplementary Materials.

\begin{table}
  \centering
  \begin{tabular}{@{}lcccc@{}}
    \toprule
    Method & PSNR $\uparrow$ & SSIM $\uparrow$ & MSE $\downarrow$ & LPIPS $\downarrow$ \\
    \midrule
    Rendered Input & 22.77 & 0.886 & 0.013 & 0.197 \\
    NHR & 25.20 & 0.906 & 0.008 & 0.209 \\
    \textit{LookinGood} & 25.13 & 0.915 & \underline{0.008} & 0.161 \\
    \midrule
    Ours (w/o detail) & 25.10 & 0.909 & 0.009 & 0.162 \\
    Ours (w/o coarse) & \underline{25.27} & \underline{0.921} & 0.010 & \underline{0.147} \\
    Ours full & \textbf{28.33} & \textbf{0.933} & \textbf{0.005} & \textbf{0.135} \\
    \bottomrule
  \end{tabular}
  \caption{Performance evaluation on test sequences against other state-of-the-art methods and the reconstructed input. The best and the second-best result for each metric are highlighted in bold and underlined respectively. The up arrow means a higher score is better and vise versa. The last three rows refer to different ablation settings of the network architecture.}
  \label{tab:test_seq}
\end{table}

\textbf{Evaluation on novel view sequences.} 
To verify the robustness of our approach when viewpoint changes, we render 8 reconstructed models of unseen subjects under a designed camera trajectory and obtain 8 novel view sequences (90 frames). Since the groundtruth is unavailable in this experiment, we introduce two popular metrics used in GAN to evaluate the quality of generated images: FID\cite{fid17} and Re-ID\cite{ReID15}, where FID evaluates the similarity between the set of real images and generated images of the same person and Re-ID evaluates the identity consistency under novel view.

\begin{figure*}[htbp]
  \centering
  \includegraphics[width=1.0\linewidth]{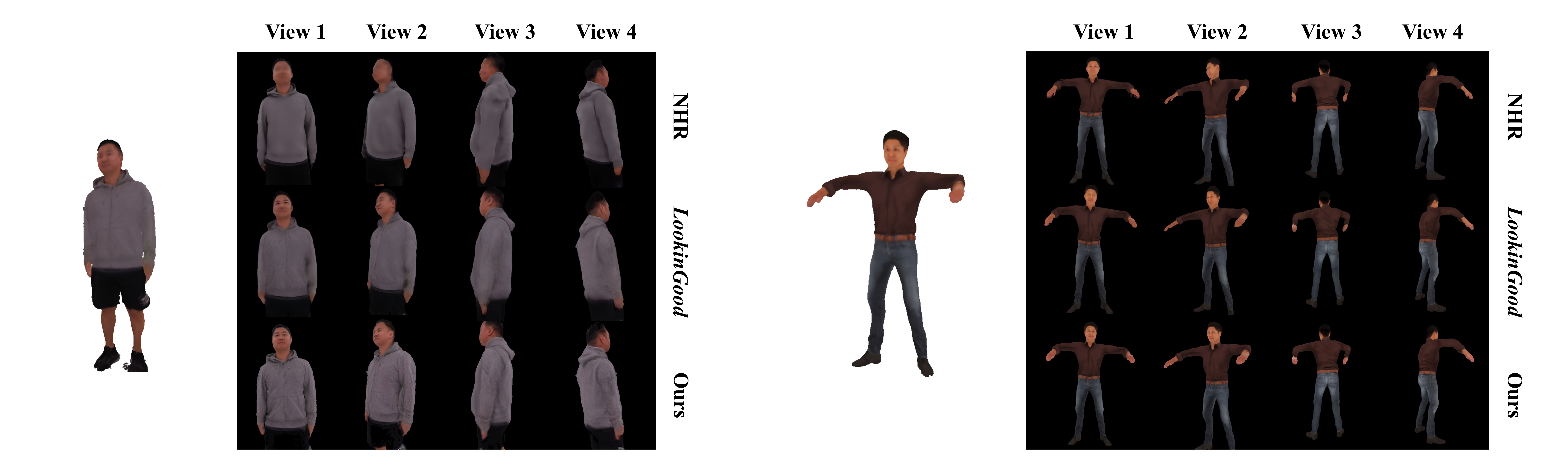}
  \caption{Results on novel view sequences. For each group of data, the first column shows the reconstructed textured geometry while other columns show results under different viewpoints. Each row corresponds to one method. Note that NHR takes a reconstructed point-cloud and a target viewpoint as input while others directly input the rendered image of the reconstructed geometry under a target viewpoint.}
  \label{fig:novel_view}
\end{figure*}

\cref{tab:novel_view} and \cref{fig:novel_view} report the quantitative and qualitative comparison against other methods under novel view. Our method performs better on both Re-ID and FID, which means the identity consistency can be well preserved while restoring enough person-specific details. As a result, although our network is trained with only sparse viewpoints, it can be robust to unseen viewpoints while continuing to produce higher-quality images than NHR and \textit{LookinGood}. 

\begin{table}
  \centering
  \begin{tabular}{@{}lcc@{}}
    \toprule
    Method & FID $\downarrow$ & Re-ID $\downarrow$ \\
    \midrule
    Rendered Input & 138.13 & 0.50 \\
    NHR & 142.26 & 0.47 \\
    \textit{LookinGood} & 138.28 & 0.48 \\
    Ours & \textbf{126.01} & \textbf{0.46}\\
    \bottomrule
  \end{tabular}
  \caption{Quality of generated images evaluation on novel view sequences against other state-of-the-art methods. FID refers to the similarity between the set of generated images and real images (unpaired data) in feature space while Re-ID refers to the identity consistency when the viewpoint changes.}
  \label{tab:novel_view}
\end{table}

\subsection{Ablation study} \label{sec:ablation}
In this section, we conduct some ablation studies to verify the availability of our proposed network architecture and analyze the selection of some significant parameters (\eg the blending ratio $\alpha$, the number of reference images $N_r$).

\textbf{Architecture.} To address the role of each branch, hence demonstrating the effectiveness of our proposed two-branch network, we use two ablation settings: (1) \textit{w/o detail}: we disable the detail branch and train the coarse branch end-to-end by supervising the $L_{c}$ loss. Note that the main difference between this setting and \textit{LookinGood} is the loss function and some details about the U-Net design as mentioned before. (2) \textit{w/o coarse}: we remove the coarse branch, which means that the warped reference features are the only input of the detail decoder, and train the detail branch with the $L_{r}$ and $L_{w}$ loss. In this setting, we do not need the rendered input images since we want to evaluate the quality of generated images without guidance from the coarse branch, thus demonstrating the superiority of our framework compared with common image warping methods.

\begin{figure}[htp]
  \centering
  \begin{subfigure}{1.0\linewidth}
    \includegraphics[width=1.0\linewidth]{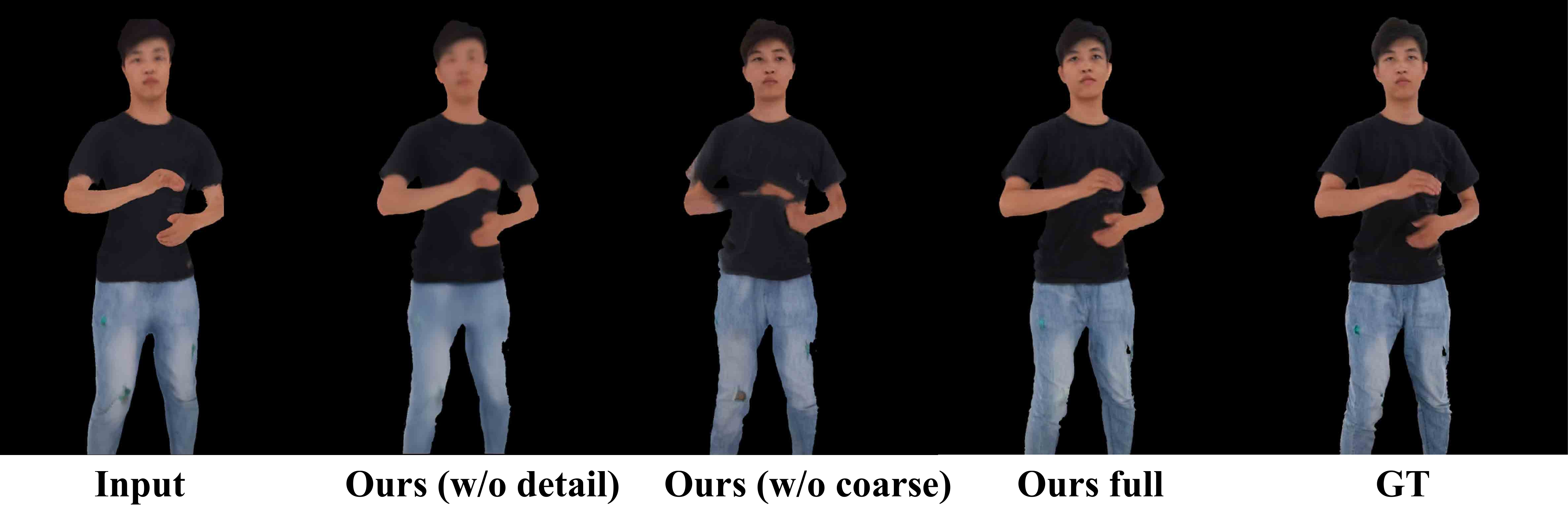}
    \caption{Network architecture.}
    \label{fig:qual_ablation_arch}
  \end{subfigure}
  \begin{subfigure}{1.0\linewidth}
    \includegraphics[width=1.0\linewidth]{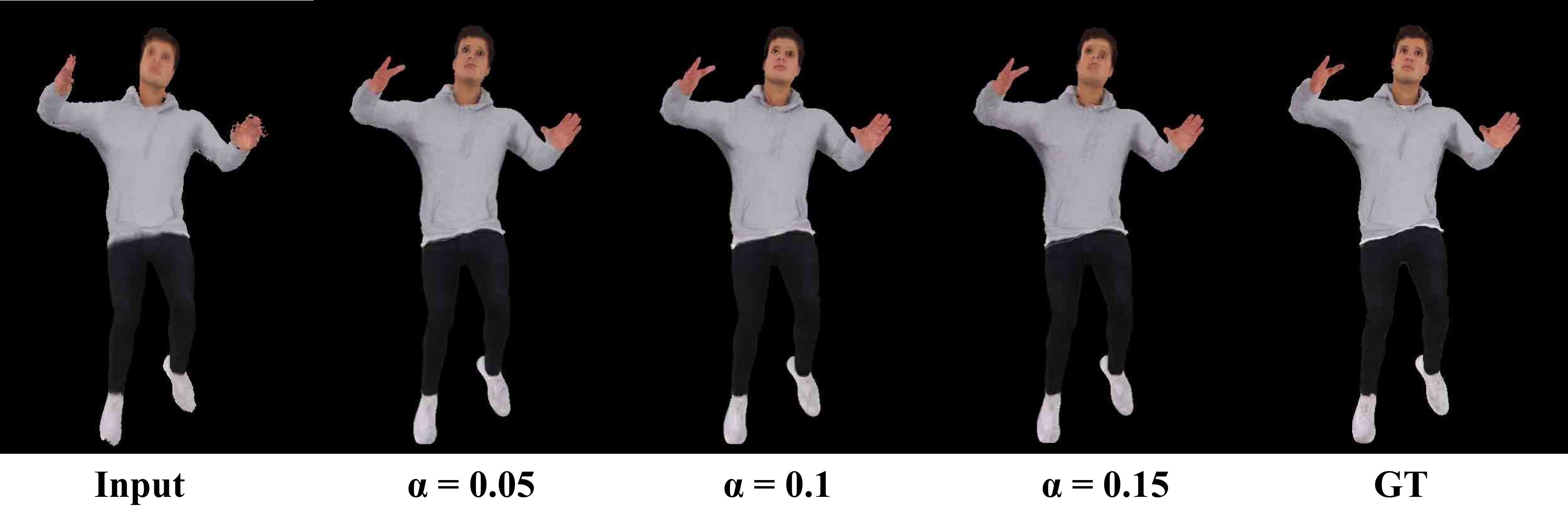}
    \caption{Blending ratio $\alpha$.}
    \label{fig:qual_ablation_alpha}
  \end{subfigure}
  \caption{Qualitative results of ablation studies on (a) different ablation settings of the network architecture and (b) different blending ratio $\alpha$ selections. Note that we only illustrate results with $\alpha$ between $0.05$ to $0.15$ since losses in this range are smaller than others, as shown in \cref{fig:quan_ablation}.}
  \label{fig:qual_ablation}
\end{figure}

We conduct the comparison experiments with the same data and metrics as \cref{sec:eval_testseq}. The quantitative and qualitative results are shown in \cref{fig:qual_ablation_arch} and \cref{tab:test_seq}. Although the coarse branch can improve the rendering quality to an extent, it will lead to loss of details. The detail branch can generate images with high-fidelity details, but it will suffer quality degradation without guidance from the coarse branch (especially in regions where warp failed). In contrast, our full model combines two branches effectively, taking advantages of both and achieving the best performance.

\textbf{Blending ratio $\alpha$.} It controls the blending ratio of the guidance features and the warped features, which is significant for improving the warping accuracy as well as predicting correct details. To find a relatively suitable $\alpha$, we sample some $\alpha$ values between 0 to 1 and draw the MSE and LPIPS loss curves under each $\alpha$ setting. Note that we calculate the average losses by using the same data as \cref{sec:eval_testseq}.

\begin{figure}[htp]
  \centering
  \includegraphics[width=1.0\linewidth]{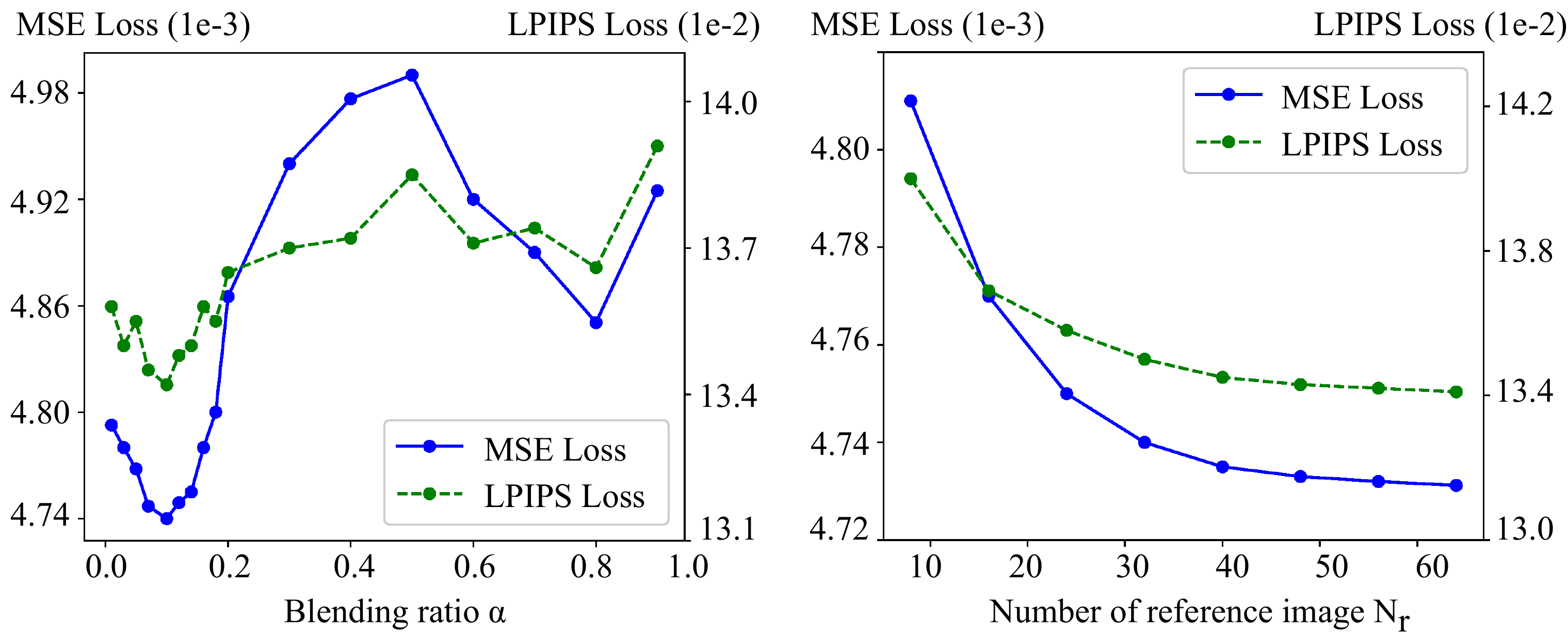}
  \caption{Quantitative results of ablation studies on the blending ratio $\alpha$ value (left) and the number of reference images $N_r$ (right). Two curves denote the average MSE and LPIPS loss respectively, which are calculated on test sequences of unseen subjects.}
  \label{fig:quan_ablation}
\end{figure}

As shown in \cref{fig:quan_ablation} and \cref{fig:qual_ablation_alpha}, when $\alpha$ is around $0.1$, our model achieves the best performance, which means that only few guidance features are needed to guide the reference image warp ``correctly" to the rendered input pose as well as guide the detail decoder to predict details in ``correct" regions.

\textbf{Number of reference images $N_r$.}
Theoretically, our performance is positively related to the number of reference images in the candidate reference set, which is illustrated in \cref{fig:quan_ablation}. To make a trade-off between computational time and performance, we set $N_r=32$ since we found less improvement when continuing to increase it.

\section{Conclusion}
We present \textit{LookinGood$^\pi$}, a real-time person-independent neural re-rendering approach for high-quality human performance capture, especially with sparse multi-view setting. Our key idea is to utilize the low-quality rendered image of reconstructed geometry as the guidance to predict person-specific details from few reference images, thus enhancing the re-rendered result. To achieve this, we design a two-branch network to integrate learning-based image warping (detail branch) into a neural re-rendering pipeline (coarse branch). We further propose a novel feature blending scheme to introduce the information of rendered image as a spatial and appearance guidance to improve both the warping accuracy and the details' fidelity. We exhaustively evaluate our approach in a reimplemented real-time performance capture system, demonstrating the state-of-the-art generalization performance on novel people. Moreover, our approach can run in real-time as well as preserve high-fidelity results under novel view, which makes it available for online VR/AR applications.

\begin{figure}
  \centering
    \includegraphics[width=1.0\linewidth]{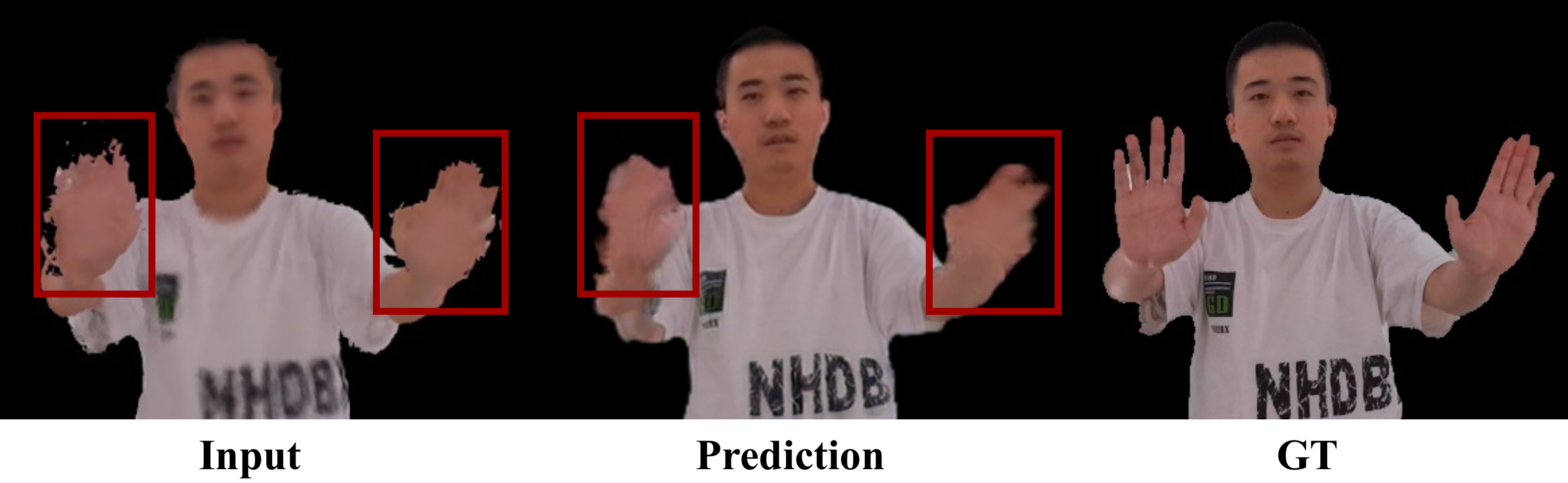}
  \caption{Limitation of our approach. When the reconstructed geometry and texture are severely corrupted in some regions due to tracking error or inaccurate depth (\eg hands), our generated result tends to be over-blurred in these regions since the guidance might be wrong.}
  \label{fig:limitation}
\end{figure}

\textbf{Limitations and future work.} Our method shares the common limitation with other RGB-D reconstruction works: the rendering result is influenced by the quality of the reconstructed geometry and texture, as shown in \cref{fig:limitation}. Moreover, similar to other segmentation-based methods, the inaccuracy of the segmentation result used for training will lead to artifacts near the boundaries. In the future, we aim to investigate a more effective scheme to learn a spatially-variant $\alpha$ to combine the guidance features and the warped reference features adaptively, which may alleviate the reconstruction error.

\renewcommand\thesection{\Alph{section}}
\renewcommand\thetable{\Alph{table}}
\renewcommand\thefigure{\Alph{figure}}
\setcounter{section}{0}
\setcounter{figure}{0}
\setcounter{table}{0}

\section*{Supplementary Material.}


\section{Evaluation on the fine-tuning data size}
As mentioned in the paper, the generalization performance of \textit{LookinGood} \cite{lookingood18} and NHR \cite{mnhr20} is highly-related to the data size used for fine-tuning. Here we provide some qualitative and quantitative experiments to illustrate and analyze this.

\begin{figure*}
  \centering
  \includegraphics[width=1.0\textwidth]{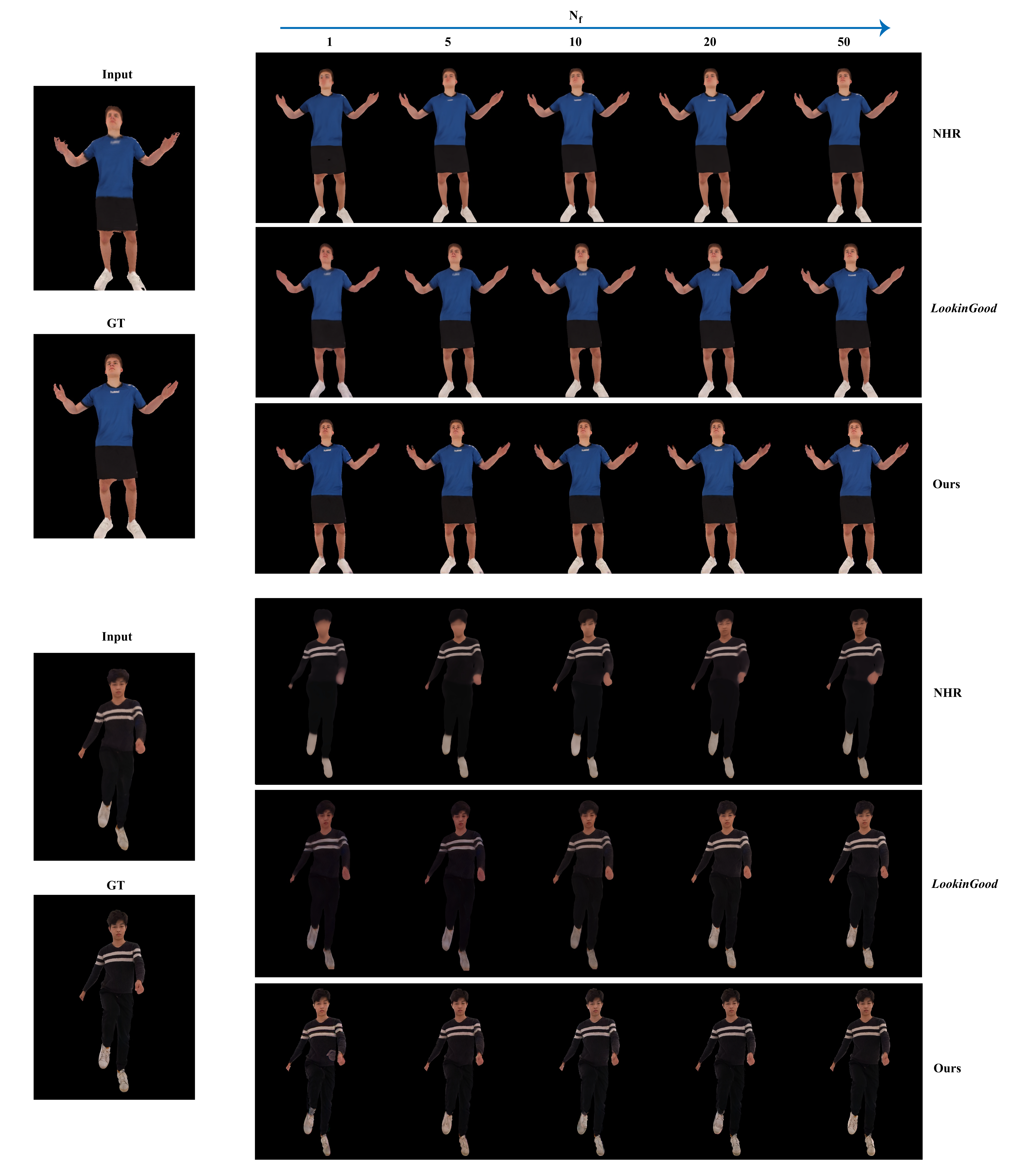}
  \caption{Qualitative evaluation on the fine-tuning data size $N_f$. For each group of data, each column corresponds to a $N_f$ setting and each row corresponds to a method. Notice that our method performs better at the beginning and converges faster than other methods. Zoom-in for the best of view.}
  \label{fig:qual_nfine}
\end{figure*}

\begin{figure}
  \centering
  \includegraphics[width=1.0\linewidth]{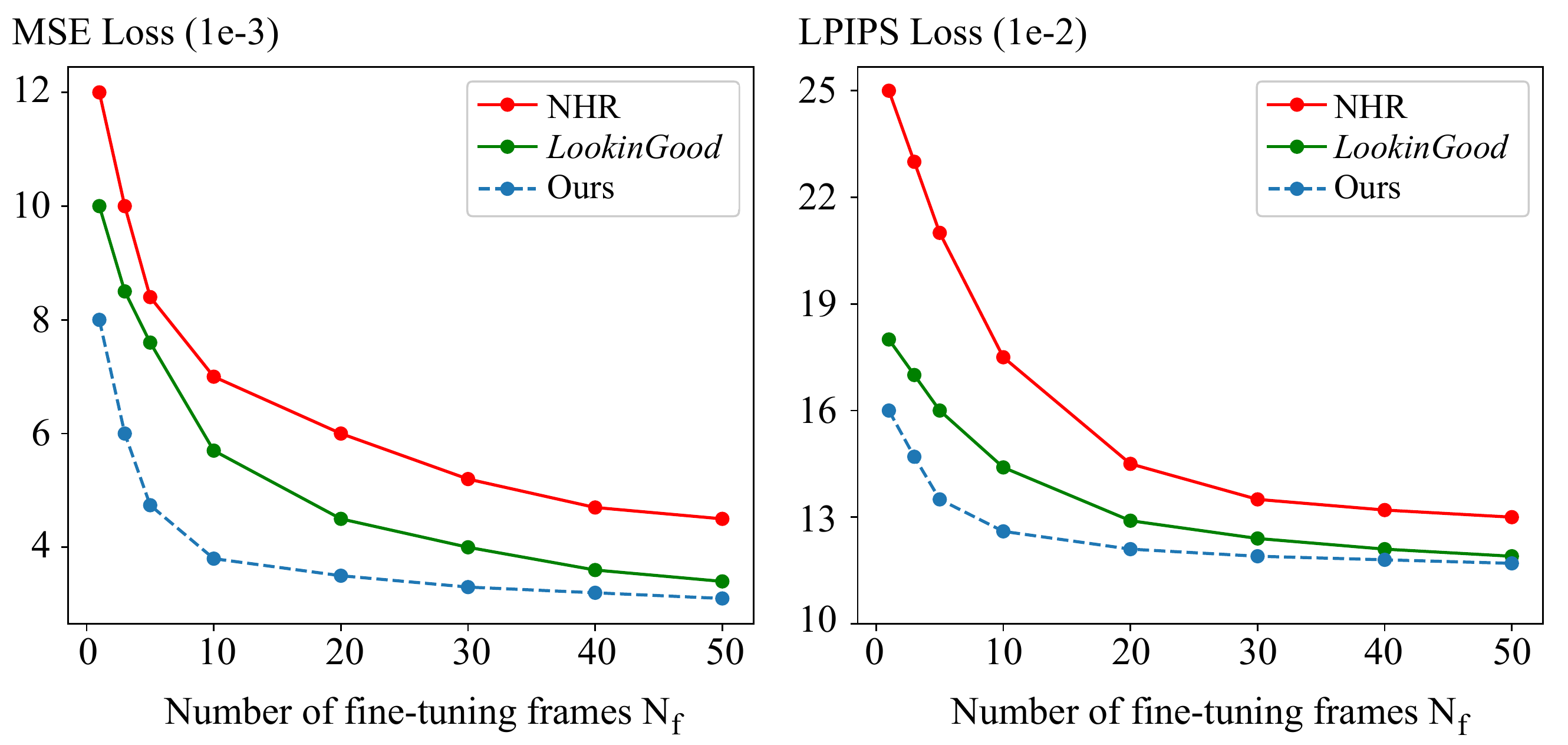}
  \caption{Quantitative results of evaluation on the fine-tuning data size setting. The curves denote the average MSE loss (left) and LPIPS loss (right) of different methods, which are calculated on test sequences of unseen subjects.}
  \label{fig:quan_nfine}
\end{figure}

We conduct the evaluation on test sequences (unseen subjects) with different fine-tuning data size (denoted by $N_f$) settings, ranged from 1 to 50. Specifically, we randomly select $N_f$ frames (each with 8 views) from the test sequences for each unseen performer. 
Then we run per-subject fine-tuning on $N_f$ frames for 20 epochs and calculate the average MSE loss and LPIPS loss on remaining frames, as shown in \cref{fig:quan_nfine}. We also provide the qualitative results in \cref{fig:qual_nfine}. 
It can be observed that as the amount of fine-tuning data increases, NHR and \textit{LookinGood} gradually achieve better generalization performance. Among them, \textit{LookinGood} obtains satisfactory result when $N_f$ increases to nearly 50, while NHR might need a larger amount of data ($i.e.$ \textgreater 50) to converge to a similar result. In contrast, our method performs better at the beginning ($i.e.\ N_f=1$) and converges fast

\section{Training details}

\textbf{Warping field training.} As described in the paper, the refine warping field $W_r$ should be trained carefully to avoid wrong gradient descent directions at the beginning of training. Specifically, in the first 15 epochs, we enable the regular term $L_w^{reg}$ and separate the image warping loss $L_w^{img}$ into two parts: (1) $L_{w_c}^{img}$ measures the warping accuracy of the coarse warping field $W_c$, (2) $L_{w_r}^{img}$ measures the warping accuracy of the whole warping field $W$ (with the refine field $W_r$). Hence, the total warp loss $L_w$ during this period is formulated as \cref{eqn:warp_warm}. 
\begin{align}\label{eqn:warp_warm}
    L_{w_c}^{img} &= |f_{w_c}(I_r)-I_{gt}| 
    \notag
    \\
    L_{w_r}^{img} &= |f_w(I_r)-I_{gt}|
    \notag
    \\
    L_w = \lambda_c^{img}*L_{w_c}^{img}&+\lambda_{r}^{img}*L_{w_r}^{img}+\lambda_w^{reg}*L_w^{reg}
\end{align}
where $f_{w_c}$ and $f_w$ denote the warping operation using the coarse field and the whole warping field respectively. $\lambda_{reg}$ is set to 1, $\lambda_c^{img}$ and $\lambda_r^{img}$ are initialized to 1 and 0 and be adjusted dynamically after 5 epochs, as defined in \cref{eqn:warp_dynamic}.
\begin{align}\label{eqn:warp_dynamic}
    \lambda_r^{img} = 0.5+0.05&*(e-5) \quad  (5\le{e}\le15)
    \notag
    \\
    \lambda_c^{img} &= 1.0-\lambda_r^{img}
\end{align}
where $e$ denotes the index of current epoch.

\textbf{Data augmentation.} To further improve the generalization ability of our approach, we adopt several data augmentation techniques in both training and fine-tuning. Specifically, for each iteration, we perform random translation, rotation, scaling on the input images before feeding to the network.

\begin{table} [htbp]
  \centering
  \resizebox{\linewidth}{!}{
    \begin{tabular}{|c|c|c|c|}
    \hline
    \multicolumn{4}{|c|}{\textbf{Network architecture}} \\
    \hline
    \multicolumn{2}{|c|}{\textbf{Coarse Branch}} & \multicolumn{2}{c|}{\textbf{Detail Branch}} \\
    \hline
    \boldsymbol{$E_c$}   & \boldsymbol{$D_c$}   & \boldsymbol{$E_r$}   & \boldsymbol{$D_d$} \\
    \hline
    \makecell[c]{$3\times3$-Conv.-$1$-$32$, \\ IN, ReLU} & Up.-$128$, SC & \makecell[c]{$7\times7$-Conv.-$3$-$32$, \\ IN, ReLU} & AB \\
    \hline
    Down.-$64$ & \makecell[c]{$3\times3$-Conv.-$1$-$128$, \\ IN, ReLU} & Down.-$64$ & \makecell[c]{Up.-$128$, AB, \\ SPADE, ReLU} \\
    \hline
    Down.-$128$ & Up.-$64$, SC & Down.-$128$ & \makecell[c]{Up.-$64$, AB, \\ SPADE, ReLU} \\
    \hline
    Down.-$256$ & \makecell[c]{$3\times3$-Conv.-$1$-$64$, \\ IN, ReLU} & Down.-$256$ & \makecell[c]{Up.-$32$, AB, \\ SPADE, ReLU} \\
    \hline
          & Up.-$32$, SC &       & $3\times3$-Conv.-$1$-$3$ \\
    \hline
          & \makecell[c]{$3\times3$-Conv.-$1$-$32$, \\ IN, ReLU} &       &  \\
    \hline
          & \makecell[c]{$3\times3$-Conv.-$1$-$4$, \\ Sigmoid} &       &  \\
    \hline
    \multicolumn{4}{|c|}{\textbf{Layer description}} \\
    \hline
    $k$$\times$$k$-Conv.-$p$-$c$ & \multicolumn{3}{c|}{\makecell[c]{Convolutional layer with kernel size $k$, stride $1$, \\ padding $p$ and output channels $c$}} \\
    \hline
    IN    & \multicolumn{3}{c|}{Instance normalization} \\
    \hline
    AvgPool-$k$ & \multicolumn{3}{c|}{Average pooling with kernel size $k$} \\
    \hline
    Down.-$c$ & \multicolumn{3}{c|}{$3\times3$ Conv.-$1$-$c$, IN, ReLU, AvgPool-$2$} \\
    \hline
    Up.-$c$ & \multicolumn{3}{c|}{$\times2$ Bilinear upsampling, $3\times3$-Conv.-$1$-$c$} \\
    \hline
    SC    & \multicolumn{3}{c|}{Skip connect (concatenate)} \\
    \hline
    AB    & \multicolumn{3}{c|}{Alpha ($\alpha$) blending} \\
    \hline
    SPADE & \multicolumn{3}{c|}{Spatially adaptive normalization layer \cite{spade19}} \\
    \hline
    \end{tabular}
    }
    \caption{Detailed architecture of our network. Conv.,Down.,Up. are the abbreviations of convolutional layer, downsampling block and upsampling block, respectively.}
  \label{tab:network_arch}%
\end{table}%

\section{Network architecture}
\cref{tab:network_arch} shows the detailed architecture of the coarse branch and the detail branch. In the following, we give a brief explanation of the implementation of each module.

\textbf{Coarse Branch.} We follow the U-Net like architecture used in \cite{lookingood18} to implement the coarse branch, which includes an encoder $E_c$ and a decoder $D_c$. To better fit with our system, we make several modifications. Firstly, we decrease the resolution of the input image by two times and upsample the predicted image to the original resolution. Secondly, we reduce the total layer number from 18 (9 for encoding and 9 for decoding) to 11 (4 for encoding and 7 for decoding). Finally, we use average pooling instead of strided convolution to perform downsampling. Similar to \cite{lookingood18}, the final layer of $D_c$ outputs a 4-channel image, including a RGB image and a foreground mask.

\textbf{Detail Branch.} The detail branch contains three key modules: the reference encoder $E_r$, the warping module $M_w$, and the detail decoder $D_d$. $E_r$ is designed in a similarly way to $E_c$. $M_w$ is implemented using the same architecture as MonkeyNet \cite{monkeynet19} (refer to their paper for the detailed architecture). Similar to MonkeyNet, the input of $M_w$ composes two parts: a reference image and two 25-channel heatmaps (converted from 25 body keypoints of the reference image and the input image respectively). Note that we downsample all the input by 4 times and obtain a downsampled warping field, which is upsampled to desired resolutions when applying. As for the detail decoder $D_d$, inspired by \cite{spade19}, we employ a SPADE layer \cite{spade19} to inject the blending features $F_b$ into $D_d$ for each layer.

\section*{ACKNOWLEDGMENTS}  Weiwei Xu is partially supported by NSFC grant No. 61732016.

{\small
\bibliographystyle{ieee_fullname}
\bibliography{egbib}
}

\end{document}